\documentclass[10pt,twocolumn,letterpaper]{article}

\usepackage[pagenumbers]{cvpr} % To force page numbers, e.g. for an arXiv version
\usepackage{multirow}
\usepackage[dvipsnames]{xcolor}

\usepackage{float}
\usepackage{pifont}
\usepackage{tabularray}
\definecolor{cvprblue}{rgb}{0.21,0.49,0.74}
\usepackage[pagebackref,breaklinks,colorlinks,citecolor=cvprblue]{hyperref}

\def\paperID{16934} % *** Enter the Paper ID here
\def\confName{CVPR}
\def\confYear{2025}

\title{Can Multimodal Large Language Models be Guided to Improve Industrial Anomaly Detection?}

\author{
Zhiling Chen$^{1}$\quad 
Hanning Chen$^{2}$\quad 
Mohsen Imani$^{2}$\quad 
Farhad Imani$^{1,}$\thanks{Corresponding author.}\\
$^1$Department of Mechanical Engineering, University of Connecticut\\
$^2$Department of Computer Science, University of California Irvine\\
{\tt\small \{zhiling.chen, farhad.imani\}@uconn.edu, \{hanningc, m.imani\}@uci.edu}
}

\begin{document}
\maketitle
\begin{abstract}
In industrial settings, the accurate detection of anomalies is essential for maintaining product quality and ensuring operational safety. Traditional industrial anomaly detection (IAD) models often struggle with flexibility and adaptability, especially in dynamic production environments where new defect types and operational changes frequently arise. Recent advancements in Multimodal Large Language Models (MLLMs) hold promise for overcoming these limitations by combining visual and textual information processing capabilities. MLLMs excel in general visual understanding due to their training on large, diverse datasets, but they lack domain-specific knowledge, such as industry-specific defect tolerance levels, which limits their effectiveness in IAD tasks. To address these challenges, we propose Echo, a novel multi-expert framework designed to enhance MLLM performance for IAD. Echo integrates four expert modules: Reference Extractor which provides a contextual baseline by retrieving similar normal images, Knowledge Guide which supplies domain-specific insights, Reasoning Expert which enables structured, stepwise reasoning for complex queries, and Decision Maker which synthesizes information from all modules to deliver precise, context-aware responses. Evaluated on the MMAD benchmark, Echo demonstrates significant improvements in adaptability, precision, and robustness, moving closer to meeting the demands of real-world industrial anomaly detection. 

\end{abstract}    
\section{Introduction}
\label{sec:intro}

In industrial inspection, ensuring product quality through efficient and accurate anomaly detection is a crucial requirement for maintaining production standards and minimizing operational risks. Traditional models for IAD typically rely on deep learning and discriminative models \cite{liu2024deep, cao2024survey, tao2022deep, chen2021surface}. While these models can identify anomalies to a certain extent, they often require extensive retraining when applied to new tasks or environments, and they struggle to adapt to the dynamic needs of industrial settings. Moreover, traditional models focus on single detection tasks without providing detailed explanations or insights, which limits their interpretability and applicability, especially when changes in production lines demand flexible and robust inspection solutions.

\begin{figure}[t]
  \centering
   \includegraphics[width=1\linewidth]{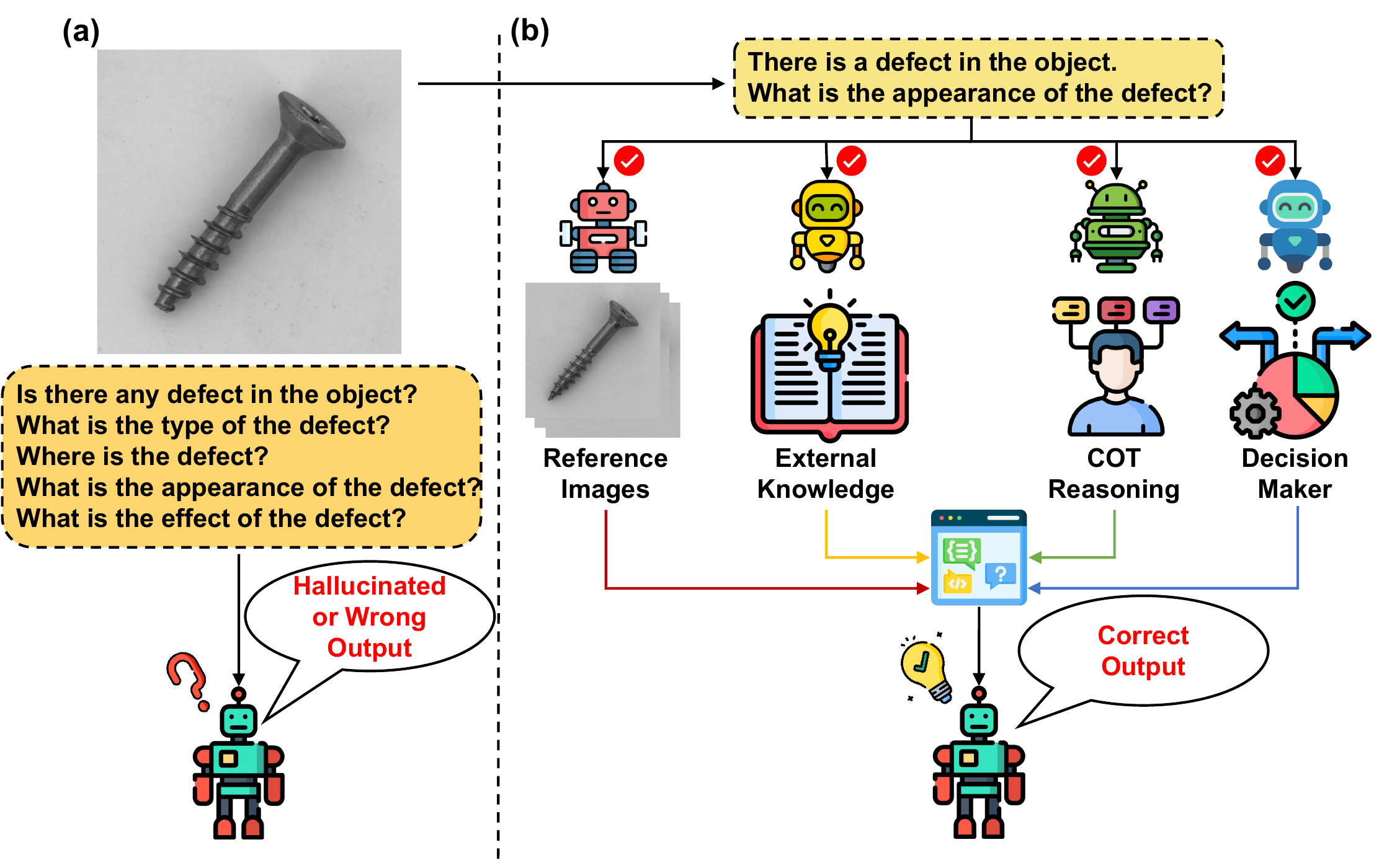}
   \caption{(a) Directly providing a query image and question to the MLLM may result in hallucinated or wrong outputs. (b) By processing the query image and question through our Echo framework, which integrates reference images, external knowledge, CoT reasoning, and a decision-making module, the system can generate accurate outputs.}
   \label{fig:introduction}
\end{figure}

Recent advancements in MLLMs hold promise for addressing some of these limitations. MLLMs demonstrate impressive generalization abilities and flexibility across various tasks by processing visual and language inputs in tandem, which allows for human-like instruction-following capabilities. 
Studies like AnomalyGPT \cite{gu2024anomalygpt} and Myriad \cite{li2023myriad} have attempted to apply MLLMs to industrial anomaly detection by training models directly on specialized datasets or customizing visual-language prompts to improve detection accuracy. However, these approaches have notable limitations. For instance, AnomalyGPT is specifically designed to recognize certain types of defects by integrating traditional IAD model outputs with large language model capabilities. Myriad, on the other hand, emphasizes the role of visual prompts tailored for industrial anomalies, aiming to enhance the model's detection accuracy across different tasks. Despite these efforts, both methods remain limited due to their reliance on narrowly defined training data and rigid output formats, which restrict the MLLM’s ability to generalize effectively across diverse industrial scenarios. 

In parallel, the overall performance of MLLMs has steadily improved with the introduction of advanced models such as LLaVA series \cite{liu2023llava, liu2023improvedllava, li2024llavaonevisioneasyvisualtask, li2024llavanextinterleavetacklingmultiimagevideo}, InternVL2 series \cite{chen2023internvl, chen2024far} and Qwen-VL \cite{Qwen-VL, Qwen2VL} series, which showcase increasingly sophisticated multimodal capabilities. These models have demonstrated notable success in general tasks requiring a combination of textual and visual understanding. However, when applied to specialized fields like IAD, challenges persist due to the unique requirements of industrial anomaly detection, such as precise feature recognition, context-specific classification, and multi-step reasoning. As shown in \cref{fig:introduction} (a), directly providing a query image and question to an MLLM can lead to hallucinated or incorrect outputs. Lacking an understanding of specific anomaly tolerance levels, the model may misinterpret the task, miss subtle defects, or generate overly general responses, limiting its effectiveness in industrial settings where precise defect identification and adherence to strict quality standards are crucial.

To systematically address these limitations and assess the effectiveness of MLLMs in IAD, the MMAD \cite{jiang2024mmad} benchmark was recently introduced. As the first comprehensive benchmark for IAD using MLLMs, MMAD evaluates a wide range of IAD tasks across diverse datasets, covering various anomaly types and object classifications. Additionally, MMAD includes two enhancement methods: Retrieval-Augmented Generation (RAG) and an Expert Agent module, which aim to expand MLLMs’ capabilities by integrating external, context-relevant knowledge and visual augmentation. Nevertheless, even with these enhancements, MLLMs still fall short of meeting industrial standards, highlighting persistent gaps in contextual understanding, adaptability, and precision.

In response to these challenges, we propose a novel multi-expert framework, Echo, designed to enhance the capabilities of MLLMs in IAD. As illustrated in \cref{fig:introduction} (b), Echo comprises four expert modules working together to produce accurate outputs: Reference Extractor, which retrieves the most similar normal image to provide a contextual reference for anomaly comparison and establish relevant anomaly tolerance levels; Knowledge Guide, which supplies task-specific domain knowledge to improve the MLLM's understanding of industry-relevant details and defect tolerances; Reasoning Expert, which enables stepwise logical reasoning to tackle complex queries; and Decision Maker, which synthesizes insights from all modules to deliver a precise and context-aware response. By processing the query image and question through these integrated modules, Echo addresses the limitations of direct MLLM querying and ensures enhanced adaptability, contextual understanding, and precision in industrial anomaly detection, with a refined sensitivity to anomaly tolerance.

To summarize, the main contributions of this paper are:
\begin{itemize}
    \item We propose \textbf{Echo}, an innovative multi-expert framework specifically designed to enhance MLLMs for industrial anomaly detection, which integrates four expert modules—Reference Extractor, Knowledge Guide, Reasoning Expert, and Decision Maker—designed to meet the nuanced requirements of fine-grained anomaly detection in industrial settings.
    
    \item By incorporating task-specific knowledge, multimodal retrieval, and structured reasoning, \textbf{Echo} enhances the accuracy and interpretability of MLLMs in complex IAD tasks.
    
    \item Through the coordinated use of expert modules, \textbf{Echo} achieves a high level of robustness and precision that better aligns with the high standards required in industrial applications.
    
    \item We conduct extensive evaluations of \textbf{Echo} on the MVTec-AD and VisA datasets within the MMAD benchmark, showing that our approach significantly outperforms existing open-source MLLMs and sets a new standard for accuracy and effectiveness in industrial anomaly detection.
\end{itemize}

\section{Related Work}
\label{sec:related work}

%-------------------------------------------------------------------------
\subsection{Multimodal Large Language Model}
Recent advancements in MLLMs have greatly expanded the capabilities of traditional LLMs by incorporating both visual and textual inputs, which has proven valuable for tasks requiring cross-modal understanding. With applications in various fields, IAD has seen the emergence of MLLMs tailored for detecting and localizing visual anomalies in industrial environments. AnomalyGPT~\cite{gu2024anomalygpt} and Myriad~\cite{li2023myriad} are among the early models that seek to improve anomaly detection by combining language processing with visual data. However, these models still face challenges in dealing with complex, unseen environments, especially when relying solely on pre-trained visual models. Moreover, most existing MLLMs rely heavily on pre-trained vision encoders, which are optimized for general-purpose tasks rather than the specific needs of industrial applications. This reliance can hinder the model's ability to detect complex, domain-specific defects in real-world industrial environments. To address these limitations, our research explores how multimodal RAG \cite{zhao2023retrieving} and CoT reasoning \cite{wei2022chain, zhang2023multimodal, gao2024cantor} can be used to guide MLLMs towards improved performance in Industrial Anomaly Detection.

%-------------------------------------------------------------------------
\subsection{Industrial Anomaly Detection}
IAD is crucial in visual inspection systems for identifying and localizing defects to ensure product quality. Traditional IAD methods primarily focus on detecting deviations by learning from normal data, which can be broadly classified into feature embedding-based and reconstruction-based approaches. Feature embedding methods model latent representations of normal samples, using distance metrics for anomaly detection, while reconstruction-based methods detect anomalies by calculating reconstruction error between input samples and their recreated versions~\cite{tax2004support, yi2020patch, zavrtanik2021reconstruction, rudolph2022fully}. Although effective, these approaches require large amounts of training data, limiting scalability in dynamic production environments.
Recent research has introduced vision-language models (VLMs) such as CLIP \cite{radford2021learning} for IAD, enabling few-shot and zero-shot detection capabilities. Approaches like \cite{jeong2023winclip}, AnomalyCLIP~\cite{zhou2023anomalyclip} and PromptAD~\cite{li2024promptad} leverage pretrained semantic understanding but are constrained by predefined anomaly concepts, limiting their generalization.

MLLMs have emerged as a potential solution for improving the flexibility and reasoning capabilities of IAD tasks. These models, which integrate both visual and textual inputs, can handle more complex and varied anomaly detection tasks. Early examples such as AnomalyGPT~\cite{gu2024anomalygpt} and Myriad~\cite{li2023myriad} have demonstrated the feasibility of training MLLMs on IAD datasets. Additionally, the latest work VMAD~\cite{deng2024vmad} integrate fine-grained visual perception with multimodal learning, achieving strong zero-shot performance on datasets like MVTecAD and VisA without requiring additional training. VMAD enhances anomaly localization and provides detailed textual explanations through mechanisms like Defect-Sensitive Structure Learning (DSSL) and Locality-Enhanced Token Compression (LTC). However, despite its effectiveness, VMAD still relies heavily on predefined concepts of anomalies and anomaly distributions, which may limit its adaptability to novel or unexpected defect types beyond those encountered in training. While its patch-based learning approach addresses some generalization challenges, the performance is constrained by the underlying assumptions about anomaly characteristics and data availability. To overcome such limitations, integrating domain knowledge and enabling more flexible, context-aware reasoning processes are essential to further improve industrial anomaly detection in real-world scenario.
% Additionally, the latest work VMAD~\cite{deng2024vmad} integrates fine-grained visual perception with multimodal learning to improve anomaly localization and explanation capabilities. However, these models are heavily dependent on predefined concepts of anomalies, restricting their capacity to adapt to novel or unexpected defect types beyond those encountered in training. 
% Furthermore, the scarcity of IAD-specific data makes these models particularly vulnerable to overfitting, which further limits their effectiveness and flexibility in practical, real-world scenarios. Therefore, incorporating additional domain knowledge, understanding defect tolerance levels, and providing guided reasoning for complex issues are essential to enhance anomaly detection capabilities.

%-------------------------------------------------------------------------
\subsection{Retrieval-Augmented Generation}

RAG \cite{lewis2020retrieval} enhances LLMs by integrating external knowledge, addressing limitations such as hallucination, improving knowledge consistency, and incorporating domain-specific information. These models retrieve relevant information from external sources in response to a query and use it to refine the model's output. 
In computer vision, retrieval-augmented methods have been applied across various contexts, including enhancing image-text alignment \cite{liu2023learning}, improving image recognition \cite{liu2024rar, hu2023reveal}, and supporting segmentation tasks \cite{liu2023learning}. These approaches demonstrate the potential of retrieval-based techniques in advancing image understanding.
However, the use of RAG for enhancing MLLMs remains relatively underexplored. While prior research has mainly focused on image or text retrieval for specific tasks \cite{chen2024mllm, yuan2024rag}, our work diverges by investigating how retrieval mechanisms can be leveraged to enhance MLLM performance in industrial anomaly detection. By incorporating image-image retrieval, image-text retrieval, and combined retrieval techniques, our approach explores how these methods can address challenges in IAD. 

\section{Method} \label{sec:method}

To overcome the challenges faced by MLLMs in visual anomaly detection, we introduce Echo, a multi-expert framework that combines a knowledge generator, a multimodal retriever, and a decision generator to enhance MLLM capabilities across diverse anomaly detection tasks. In \cref{Preliminaries}, we outline the overall architecture of Echo. We then present a detailed description of each core component: Knowledge Generation (\cref{Knowledge Generation}), Multimodal Retriever (\cref{Multimodal Retriever}), and Decision Generation (\cref{Decision Generation}).

\begin{figure*}
  \centering
  \includegraphics[width=1\linewidth]{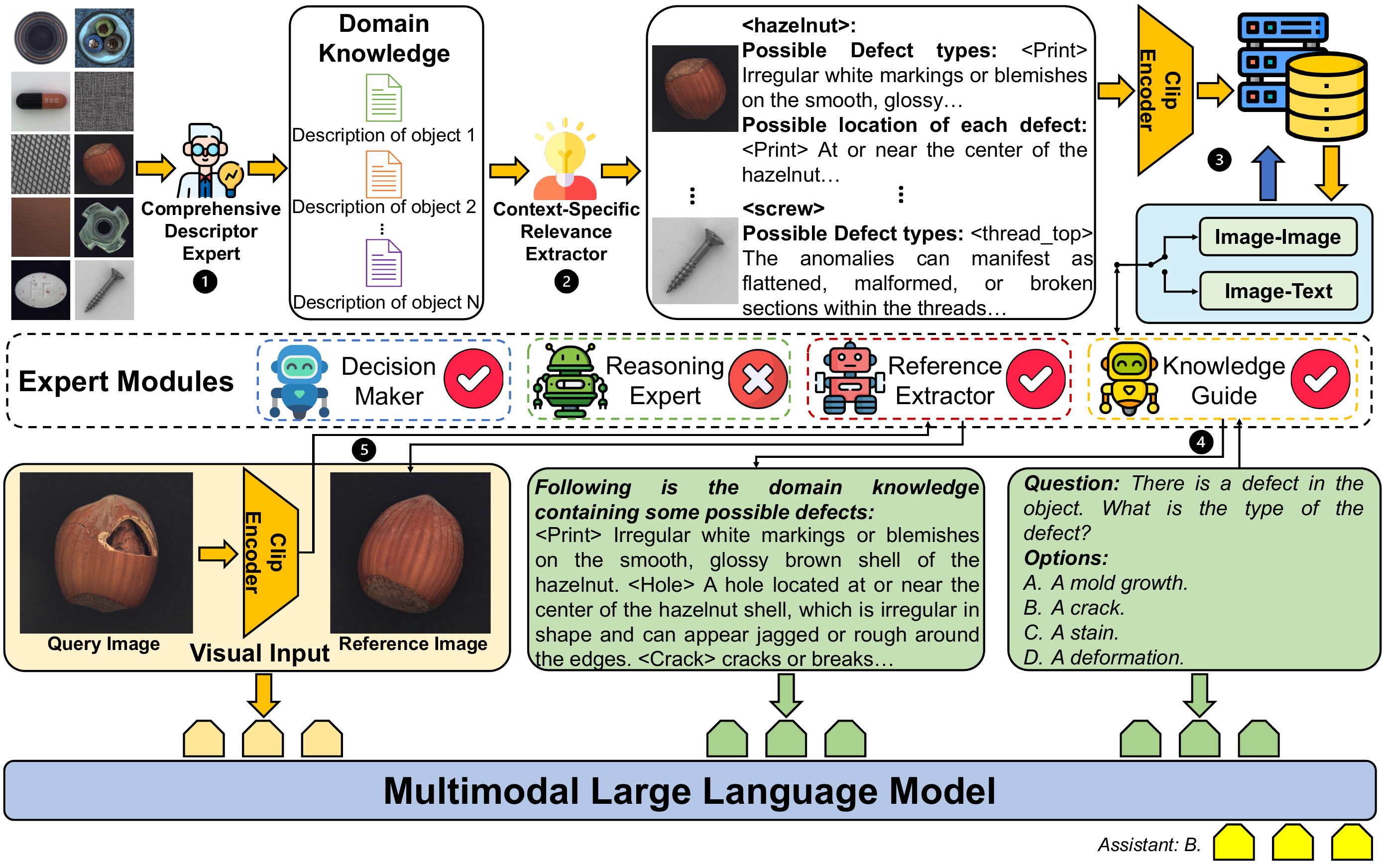}
  \caption{Framework of Echo. The Echo framework uses multiple expert modules to enhance industrial anomaly detection in MLLMs. A query image and question are processed by a CLIP encoder, with the Reference Extractor retrieving a similar normal image and the Knowledge Guide gathering relevant defect information. These inputs are fed into the MLLM, where the Reasoning Expert aids complex inference, and the Decision Maker combines outputs to generate a precise, context-aware response.}
  \label{fig:model}
\end{figure*}

\subsection{Preliminaries} \label{Preliminaries}
Echo consists of three stages: Knowledge Generation, Multimodal Retrieval, and Decision Generation, as is shown in \cref{fig:model}. In the Knowledge Generation stage, Echo extracts domain-specific information crucial for accurate anomaly detection across various industrial standards. This stage combines automated methods with manual review to ensure precise and context-specific knowledge for each query. In the Multimodal Retrieval stage, the system retrieves relevant information from an extensive memory of multimodal embeddings.
During the Decision Generation stage, Echo's input consists of $X = \{ I_q, T_q \}$, where $I_q$ denotes the query image and $T_q$ represents the question text. Based on $I_q$, Echo retrieves the most similar normal reference image, $I_r$, and gathers context-specific knowledge, $T_k$, related to the relevant object. Depending on the question type, a prompt $P$ may be added to guide the decision generation. Formally, given an input query $X$, a decision $D$ is generated as follows: $D = F(X)$, where $F$ denotes the decision generator (an MLLM).

\subsection{Knowledge Generation} \label{Knowledge Generation}
In industrial applications, standards and tolerance thresholds for defects vary widely across sectors, necessitating context-specific knowledge for accurate anomaly detection. Our framework leverages domain knowledge from MMAD~\cite{jiang2024mmad}, which provides comprehensive descriptions of potential defects and exemplar good samples for each object, as shown in \cref{fig:model} \ding{182}. However, different queries may require different subsets of this knowledge, and excessive or irrelevant information could interfere with the model’s judgment. To address this potential problem, Echo employs a Context-Specific Relevance Extractor that dynamically tailors the domain knowledge to the specific requirements of each query, filtering and refining relevant details based on the query’s focus—such as defect type, location, or potential impact, as is shown in \cref{fig:model} \ding{183}. Powered by GPT-4o, this extractor enhances the contextualization of defect attributes, ensuring relevance and precision, while manual review adds an extra layer of accuracy, capturing subtle distinctions often overlooked by automated methods. Through this structured and adaptable knowledge generation process, Echo provides context-aware anomaly detection, supporting diverse industrial applications with tailored, high-precision insights.

\begin{figure*}[]
  \centering
  \includegraphics[width=1\linewidth]{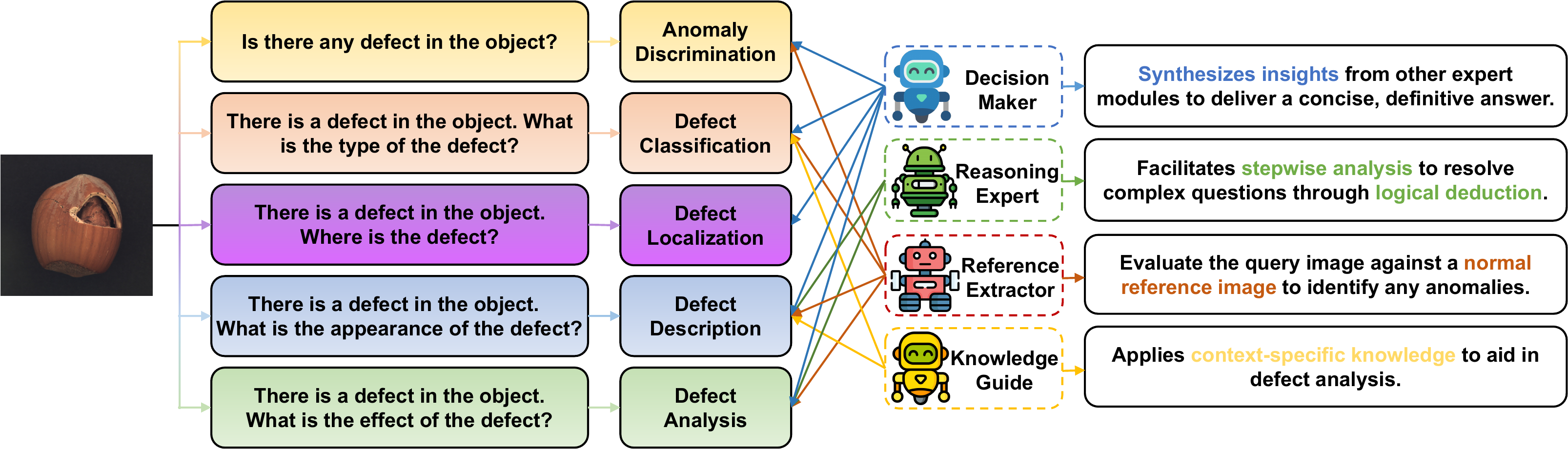}
  \caption{Experts Module. The Experts Module in Echo assigns specific modules to address different types of industrial anomaly detection queries, activating the most relevant experts based on query requirements to deliver precise, context-aware responses.}
  \label{fig:experts}
\end{figure*}

\subsection{Multimodal Retriever} \label{Multimodal Retriever}
The multimodal retriever plays a crucial role in querying an extensive multimodal external memory or database to retrieve information relevant to the given input query or context. A significant challenge in the multimodal retrieval process is the efficient encoding and storage of a large volume of image and text embeddings, which is essential to enable quick and accurate retrieval.

\noindent\textbf{Extracting the Multi-modal Embeddings.}
We utilize the CLIP model to extract the multi-modal embeddings, as is shown in \cref{fig:model} \ding{184}. Given a data sample $(x_i,c_i)$ from the dataset $D$ containing the image $x_i$ and class name $c_i$, we use the CLIP image encoder $\phi_{\text{img}}$ to extract the image embedding $e_{\text{img}}$ $\in$ $\mathbb{R}^d$ and the CLIP text encoder $\phi_{\text{text}}$ to extract the text embedding $e_{\text{text}}$ $\in$ $\mathbb{R}^d$. The symbol $d$ indicates the feature dimension (e.g., d=576 for CLIP ViT-B/16). These image and text embeddings are stored in the memory $\mathcal{M}$ to enable efficient retrieval. 

\noindent\textbf{Fast Retrieval Optimization.}
The retriever is typically designed using an exhaustive search approach, where similarity scores (e.g., cosine similarity) are computed for each vector stored in memory $\mathcal{M}$, followed by selecting the top-$k$ matches. However, as the dataset size increases to millions of embeddings, the efficiency of this brute force search diminishes significantly. To address this, we employ an indexing strategy utilizing the HNSW (Hierarchical Navigable Small World) algorithm~\cite{malkov2018efficient}. HNSW reduces dimensionality, enabling the construction of a more compact and efficient index. Specifically, HNSW organizes vectors from the original $\mathbb{R}^d$ space into a hierarchical graph structure, enabling efficient similarity comparisons and significantly accelerating the retrieval process without requiring explicit dimensionality reduction.

\subsection{Decision Generation} \label{Decision Generation}
After successfully constructing memory $M$ by using our multimodal retriever, our next step is to integrate the retrieval process and leverage MLLMs to enhance the performance in anomaly detection tasks. This involves generating a decision $D$ that considers and addresses the question. Below, we provide a detailed introduction to the Decision Generation process of Echo:

\noindent\textbf{Decision Planning.} Echo initiates the decision generation process by acting as a high-level decision planner, wherein it interprets the input query $X = \{ I_q, T_q \}$. This phase begins by classifying the question type to determine the necessary expert modules for processing.

\noindent\textbf{Expert Modules Unveiled.} As shown in the \cref{fig:experts}, we provided detailed information on the characteristics of each expert module for Echo, with the aim to allocate tasks to each expert module based on the specific requirements of the problem, as follows: 
\textit{Knowledge Guide:} This module integrates context-specific knowledge to inform the analysis and interpretation of detected defects, as discussed in \cref{Knowledge Generation}. As shown in \cref{fig:model}  \ding{185}, when the model receives a question like “What is the type of the defect?”, the Knowledge Guide retrieves context-specific information from the memory $\mathcal{M}$ based on the query image and question, ensuring that Echo’s responses are aligned with regulatory and practical standards in the industrial domain.
\textit{Reference Extractor:} This module evaluates the query image by comparing it against a repository of normal, defect-free images to identify subtle deviations that may indicate anomalies. As shown in \cref{fig:model} \ding{185}, when the model receives a query image and question, it uses the same CLIP encoder to extract the query image embedding and retrieves the most similar normal image from the memory $\mathcal{M}$ based on cosine similarity, providing a reference for defect tolerance assessment to the MLLM.
\textit{Reasoning Expert:} This module facilitates logical deduction and structured analysis to resolve complex questions that require multi-step reasoning, such as "What is the likely cause of the defect?", which allows Echo to approach each question methodically, simulating the cognitive processes of a skilled inspector by synthesizing information from various sources within the system. 
\textit{Decision Maker:} This module synthesizes insights from other expert modules to make a final, clear decision on the correct answer. For questions such as “Which option best describes the defect in the object?”, the Decision Maker integrates information from reference samples, external knowledge, and detailed observations to deliver a concise and definitive response.

\noindent\textbf{Expert Module Selection.} After identifying the type of query through the Decision Planning stage, Echo activates the most relevant expert modules to address the specific requirements of each query. This selection process is guided by a pre-defined mapping that associates query types with the modules best suited for providing accurate and contextually appropriate responses, as shown in \cref{fig:experts}. For example, general queries such as “Is there any defect in the object?” are handled by the \textit{Reference Extractor} module and \textit{Decision Maker} module, which work together by comparing the reference image with the query image to identify discrepancies and make a final decision. 
For queries requiring additional external information, such as “What is the type of the defect?”, Echo combines outputs from the \textit{Reference Extractor}, \textit{Knowledge Guide} and \textit{Decision Maker}, which consolidates information from multiple modules to simulate the reasoning processes of a skilled inspector.
Conversely, more complex queries that involve defect description, such as “What is the appearance of the defect?”, require the activation of multiple modules in tandem. In these cases, Echo first employs the \textit{Reference Extractor} to compare the query image with a set of normal samples, establishing a baseline; it then invokes the \textit{Knowledge Guide} to provide domain-specific insights into the implications of detected deviations, with the \textit{Reasoning Expert} guiding the inference process. Finally, the \textit{Decision Maker} integrates insights from all experts, ensuring Echo’s response is thorough and contextually aligned with industrial standards.

\noindent\textbf{Multi-Expert Output Synthesis.} Once the selected expert modules have completed their tasks, Echo synthesizes their outputs to deliver a unified response that captures both visual and contextual aspects of the detected anomaly. For example, in response to queries such as “What is the effect of the defect?”, Echo combines insights from the \textit{Knowledge Guide} with contextual information to deliver a complete assessment, encompassing defect identification and evaluating its potential impact according to regulatory standards. When handling queries that involve comparative analysis, such as those related to defect appearance, Echo reconciles outputs from the \textit{Reference Extractor} and the \textit{Knowledge Guide}, with inference led by the \textit{Reasoning Expert}, and makes a final decision through the \textit{Decision Maker} to provide a response that is both visually grounded and contextually relevant.

The above components are integrated to form the Echo's final decision generation prompt $P$, which combined with visual input $I$ and text input $T$, constitutes the complete input for the prompt generation stage of Echo, prompting Echo to deliver a deliberate decision $D$. This decision generation method is a core contribution of our work, where the MLLM serves as the central decision generator, orchestrating a suite of expert modules that each contribute specialized capabilities, akin to the functions of distinct sensory and analytical systems. Through dynamic module selection and prompt generation, Echo tailors responses to diverse types of queries, enhancing efficiency and accuracy across various anomaly detection tasks. By incorporating in-context examples and leveraging multi-modal retrieval, Echo not only learns to adapt but also maintains consistency with industry standards, significantly reducing the likelihood of generating hallucinated or inconsistent outputs.

\section{Experiments} \label{sec:experiment}

\subsection{Experimental Settings}

\textbf{Datasets.} In this study, we conduct comprehensive experiments using two datasets from the MMAD \cite{jiang2024mmad} benchmark: MVTec-AD \cite{bergmann2019mvtec} and VisA \cite{zou2022spot}. These datasets were selected based on their relevance to anomaly detection tasks and their representation of diverse industrial objects and defect types. MVTec-AD contains 15 categories of industrial objects with 700 $\times$ 700 to 1024 $\times$ 1024 resolutions per image and VisA contains 12 categories of industrial objects with approximately 1500 $\times$ 1000 resolutions per image. 

\begin{figure}[t]
  \centering
   \includegraphics[width=1\linewidth]{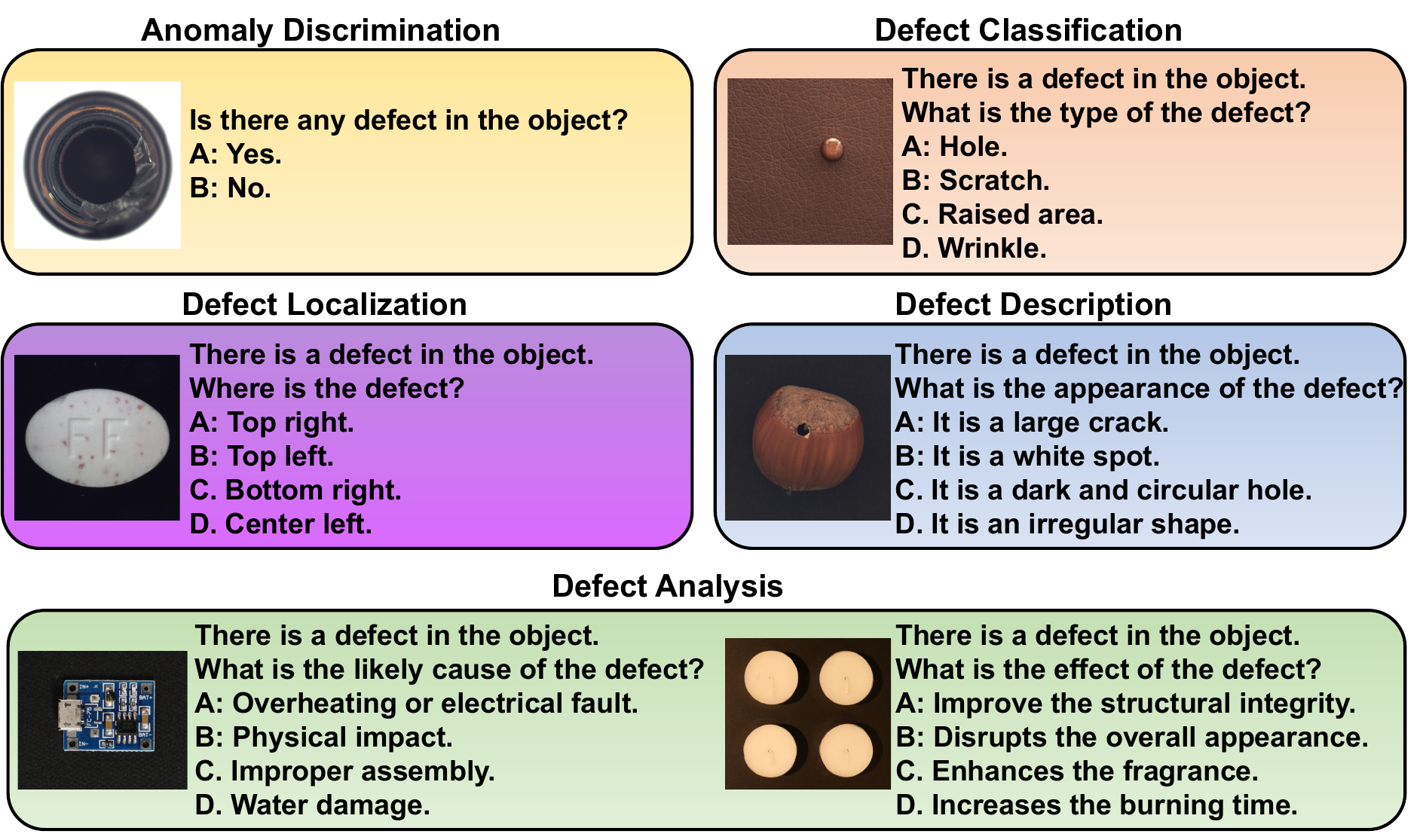}
   \caption{Example of 5 subtasks from MMAD, each presented in a multiple-choice format with distractor options designed to test the model's anomaly detection capabilities.}
   \label{fig:data}
\end{figure}

To align the experimental design with our primary research objective of evaluating the anomaly detection capabilities of MLLMs, we removed questions related to object classification and object analysis from the seven subtasks in the MMAD benchmark. As shown in \cref{fig:data}, the subtasks that were retained pertain to anomaly discrimination, defect classification, defect localization, defect description, and defect analysis, as these are directly relevant to assessing industrial anomaly detection models.

\begin{table*}[t]
\centering
\caption{Accuracy score (\%) on MVTec-AD and VisA for five tasks, which includes Anomaly Discrimination, Defect Classification, Defect Localization, Defect Description and Defect Analysis.}
\resizebox{\textwidth}{!}{
\begin{tabular}{lcccccccccccc}
\hline
\multirow{2}{*}{\textbf{Model}} & \multirow{2}{*}{\textbf{Scale}} & \multicolumn{5}{c}{\textbf{MVTec-AD}}                                                                                & \multicolumn{5}{c}{\textbf{VisA}}                                                                                    & \multirow{2}{*}{\textbf{Average}} \\ \cline{3-12}
                                &                                 & \textbf{Discrimination} & \textbf{Classification} & \textbf{Localization} & \textbf{Description} & \textbf{Analysis} & \textbf{Discrimination} & \textbf{Classification} & \textbf{Localization} & \textbf{Description} & \textbf{Analysis} &                                   \\ \hline
Random Chance                   & -                               & 50.00                   & 25.00                   & 25.00                 & 25.00                & 25.00             & 50.00                   & 25.00                   & 25.00                 & 25.00                & 25.00             & 30.00                             \\ \hline
InternVL2                       & 4B                              & 70.96                   & 44.81                   & 66.97                 & 59.52                & 87.39             & 63.29                   & 20.17                   & 53.71                 & 58.15                & 70.25             & 59.52                             \\
InternVL2                       & 8B                              & 76.88                   & 51.04                   & 59.18                 & 64.55                & 85.73             & 68.85                   & 35.04                   & 55.81                 & 59.24                & 75.24             & 63.16                             \\
MiniCPM-V2.6                    & 8B                              & 72.50                   & 64.07                   & 68.65                 & 79.55                & 90.04             & 64.41                   & 50.67                   & 57.73                 & 69.50                & 68.44             & 68.56                             \\
LLaVA-NeXT                      & 7B                              & 78.42                   & 45.23                   & 64.96                 & 68.18                & 87.14             & 56.98                   & 40.08                   & 58.90                 & 62.86                & 68.87             & 63.16                             \\
LLaVA-OneVision                 & 7B                              & \textbf{94.09}                  & \textbf{79.59}                   & \textbf{78.12}                 & 83.18                & 91.29             & 76.46                   & 52.44                   & 60.40                 & 68.49                & 75.92             & 76.00                             \\
Qwen2-VL                        & 2B                              & 73.21                   & 60.08                   & 65.46                 & 74.86                & 88.63             & 59.79                   & 37.31                   & 58.15                 & 66.81                & 68.53             & 65.28                             \\ \hline
Qwen2-VL                        & 7B                              & 82.26                   & 68.46                   & 76.11                 & 82.19                & 92.28             & 73.00                   & 57.06                   & \textbf{63.41}                 & \textbf{74.71}                & 77.82             & 74.73                             \\
Qwen2-VL (+Echo)                & 7B                              & 89.65 (\textcolor{red}{+7.39})           & 72.86 (\textcolor{red}{+4.40})           & 76.11 (=)             & \textbf{85.57} (\textcolor{red}{+3.38})        & \textbf{93.20} (\textcolor{red}{+0.92})     & \textbf{76.79} (\textcolor{red}{+6.79})           & \textbf{62.35} (\textcolor{red}{+5.29})           & \textbf{63.41} (=)             & 74.03 (\textcolor{green}{-0.68})        & \textbf{79.11} (\textcolor{red}{+1.29})      & \textbf{77.31} (\textcolor{red}{+2.58})                     \\ \hline
\end{tabular}}
\label{main result}
\end{table*}

\noindent\textbf{Implementation Details} We utilize a frozen CLIP ViT B/16 model as the visual encoder to process input image and extract the corresponding image embeddings. For the retrieval phase, we perform a search over the stored embeddings in memory $M$ using the Hierarchical Navigable Small World (HNSW) algorithm \cite{malkov2018efficientrobustapproximatenearest}. The vector database is implemented using Faiss \cite{johnson2019billion}, enabling efficient large-scale similarity searches. We select MLLMs including InternVL2 \cite{chen2024far}, MiniCPM-V2.6 \cite{yao2024minicpm}, LLaVA-NeXT \cite{li2024llavanextinterleavetacklingmultiimagevideo}, LLaVA-OneVision \cite{li2024llavaonevisioneasyvisualtask}, Qwen2-VL \cite{Qwen2VL} as baselines.

\subsection{Main Results}

We compare the performance of our proposed model, Qwen2-VL (+Echo), with various baseline models across the MVTec-AD and VisA datasets, as shown in \cref{main result}. Each model's performance is evaluated across multiple tasks, including Discrimination, Classification, Localization, Description, and Analysis.
Our results demonstrate that Qwen2-VL (+Echo) achieves state-of-the-art performance, surpassing other models in key metrics, particularly in Discrimination and Classification tasks on both datasets. Specifically, compared to the baseline Qwen2-VL, the integration of Echo provides consistent improvements, with gains of +7.39\% in Discrimination and +4.40\% in Classification on MVTec-AD, and +6.79\% in Discrimination and +5.29\% in Classification on VisA. 
Compared to other large-scale models, Qwen2-VL (+Echo) demonstrates consistently superior performance, particularly in tasks that require high interpretability and nuanced visual understanding. For example, on the MVTec-AD dataset, our model achieves 89.65\% in Discrimination and 82.57\% in Analysis, representing a substantial improvement over other state-of-the-art models, such as LLaVA-OneVision, across both MVTec-AD and VisA. The addition of Echo yields an average performance increase of +2.58\% across all tasks on both datasets, underscoring its effectiveness in enhancing the model’s interpretative capabilities and overall performance in complex industrial applications.
These findings underscore two significant insights: (1) the integration of Echo into MLLM enhances both the interpretive and discriminative capabilities of the model, yielding state-of-the-art results across complex anomaly detection tasks, and (2) MLLM equipped with structured expert modules like Echo can achieve a more nuanced understanding of defect characteristics, thus addressing both general and specific task requirements within industrial anomaly detection.

\subsection{Q\&A, Multiple-choice or True/False?}

\begin{figure}[t]
  \centering
   \includegraphics[width=1\linewidth]{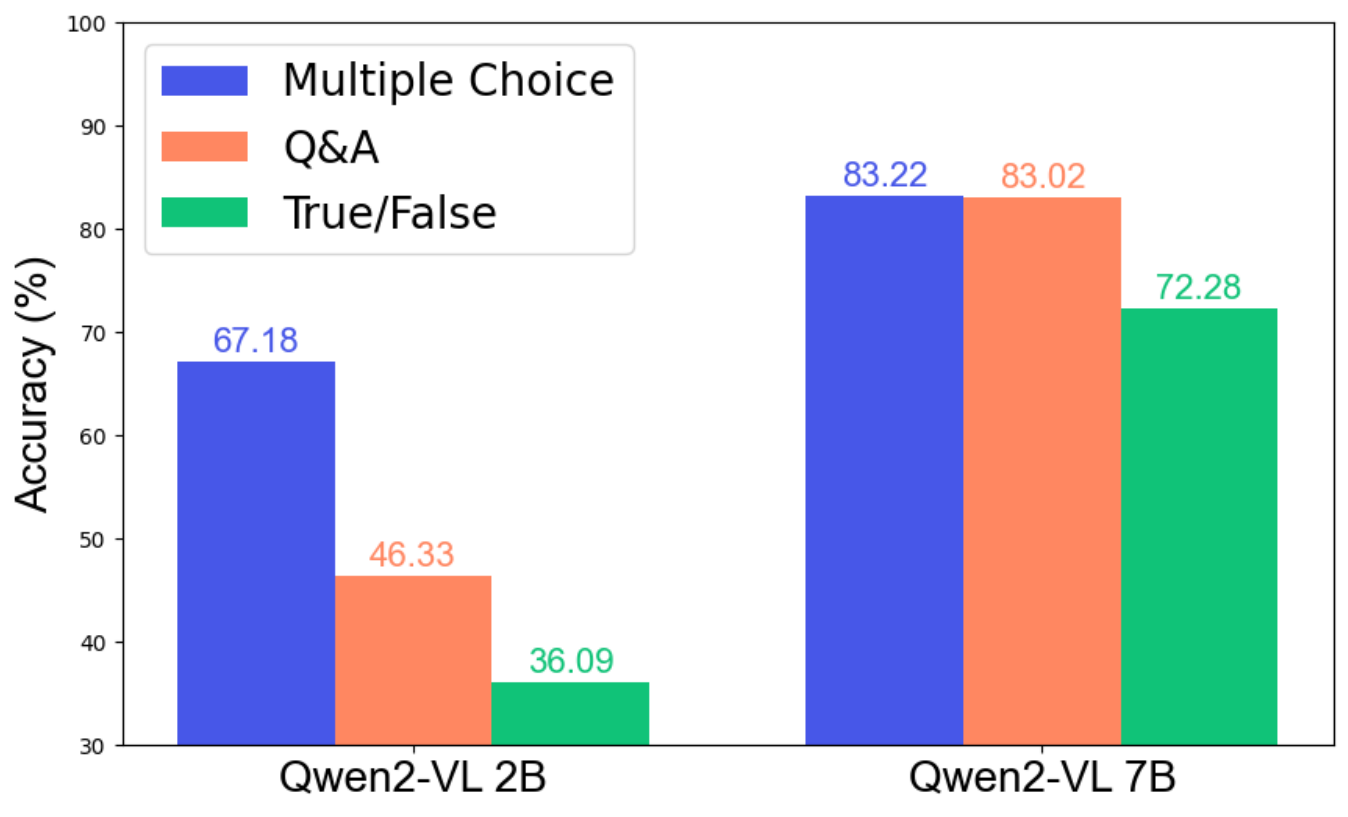}
   \caption{Multiple Choice, Q$\&$A and True/False evaluation on Anomaly Discrimination.}
   \label{fig:question type}
\end{figure}

We further investigate the impact of different prompt formats on Anomaly Discrimination. We primarily focus on three prompt modes: Q$\&$A, multiple-choice, and True/False. The multiple-choice mode is the default mode for the baseline models. In the Q$\&$A format, the model is asked open-ended questions without any answer choices. This format requires the model to independently determine and respond if anomalies are present. In contrast, the True/False format presents a specific statement like “The query image contains no anomalies or defects compared to the normal sample,” prompting the model to judge if the statement is true or false. Taking the Qwen2-VL 2B and 7B as examples, the experimental results are shown in \cref{fig:question type}. The results show that the multiple-choice format consistently outperforms both Q$\&$A and True/False formats in anomaly discrimination accuracy across the Qwen2-VL 2B and 7B models. While the Q$\&$A format also performs relatively well, allowing the model to assess anomalies with a degree of interpretative freedom, the True/False format lags behind, especially in the smaller Qwen2-VL 2B model. We suspect that this discrepancy may be due to the tendency of MLLMs to exhibit sycophantic behavior \cite{wei2023simple}. In conclusion, these findings suggest that the multiple-choice mode is the most effective format for enhancing the accuracy in anomaly detection tasks.

\begin{table}[t]
\centering
\caption{The impact of different levels of external knowledge on defect classification and defect description. Here CSK denotes context-specific knowledge, DK denotes domain knowledge, EK denotes external knowledge.}
\resizebox{\columnwidth}{!}{%
\begin{tabular}{llccc}
\hline
\textbf{Model}               & \textbf{Method} & \textbf{Defect Classification} & \textbf{Defect Description} & \textbf{Mean} \\ \hline
\multirow{3}{*}{Qwen2-VL 2B} & w/o EK          & 44.82                          & 67.21                       & 56.02         \\
                             & + DK            & 45.10 (\textcolor{red}{+0.28})                  & \textbf{70.70} (\textcolor{red}{+3.49})               & 57.90 (\textcolor{red}{+1.88}) \\
                             & + CSK           & \textbf{46.87} (\textcolor{red}{+2.05})                  & 70.07 (\textcolor{red}{+2.86})               & \textbf{58.47} (\textcolor{red}{+2.45}) \\ \hline
\multirow{3}{*}{Qwen2-VL 7B} & w/o EK          & 63.55                          & 78.75                       & 71.15         \\
                             & + DK            & 65.98 (\textcolor{red}{+2.43})                  & 79.31 (\textcolor{red}{+0.56})               & 72.65 (\textcolor{red}{+1.50}) \\
                             & + CSK           & \textbf{67.61} (\textcolor{red}{+4.06})                  & \textbf{79.80} (\textcolor{red}{+1.05})               & \textbf{73.71} (\textcolor{red}{+2.56}) \\ \hline
\end{tabular}
}
\label{exp: external knowledge}
\end{table}

\subsection{Does context-specific knowledge better than domain knowledge?}

To evaluate the effectiveness of context-specific knowledge versus complete domain knowledge, we conducted experiments using the Qwen2-VL models with both 2B and 7B parameter configurations. For each model, we tested three configurations: (1) with context-specific knowledge, (2) with domain knowledge, and (3) without any external knowledge. As shown in \cref{exp: external knowledge}, for both the 2B and 7B models, incorporating context-specific knowledge resulted in improved performance across defect classification and defect description tasks. Specifically, the Qwen2-VL 7B model with context-specific knowledge achieved the highest mean accuracy of 73.71\%, outperforming the domain knowledge and no external knowledge settings. This demonstrates that tailoring knowledge to specific questions can enhance the model's ability to classify and describe defects accurately. The results also indicate that redundant knowledge, such as general domain information not directly relevant to the specific question, can introduce noise and potentially hinder the model’s decision-making process. The consistent improvement observed in the context-specific knowledge setting highlights its potential utility in applications where precise, context-aware information is critical, minimizing unnecessary information that may otherwise confuse the model.

\begin{figure}[t]
  \centering
   \includegraphics[width=1\linewidth]{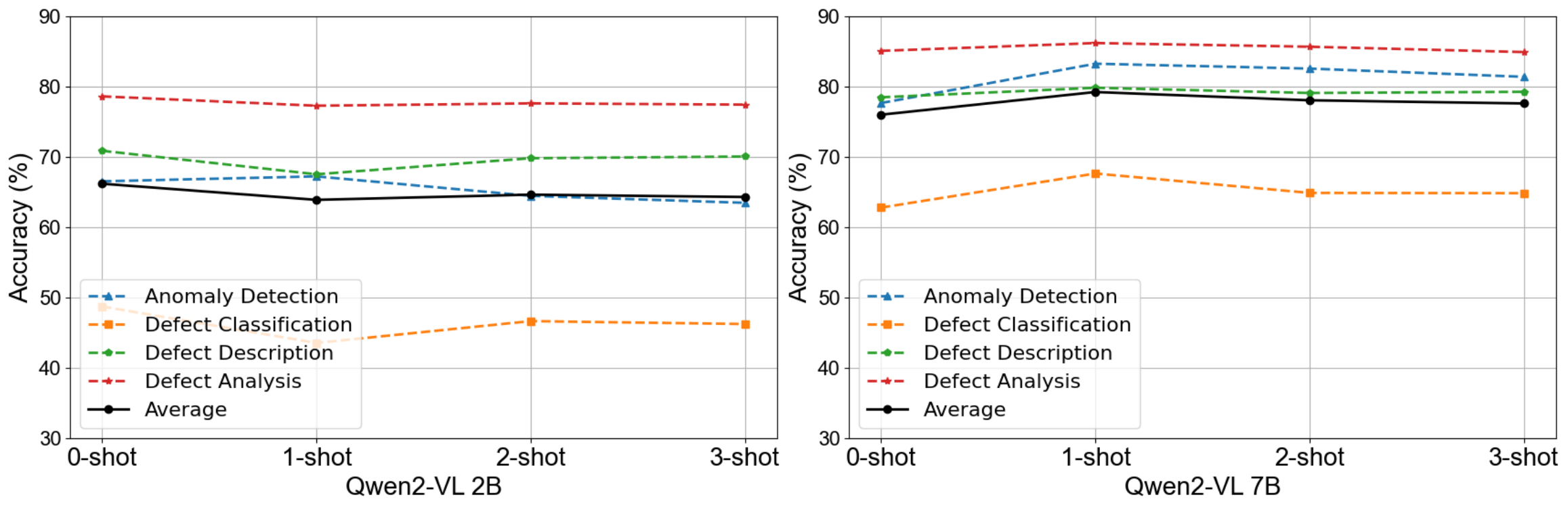}
   \caption{Performance trends across different counts of normal samples.}
   \label{fig:shots}
\end{figure}

\subsection{Can reference normal sample help anomaly detection?}

To investigate whether reference normal samples can help anomaly detection, we conducted experiments with varying numbers of reference samples (0-shot, 1-shot, 2-shot, and 3-shot) across tasks such as Anomaly Detection, Defect Classification, Defect Description, and Defect Analysis, using the Qwen2-VL 2B and Qwen2-VL 7B models. As shown in \cref{fig:shots}, incorporating a single reference sample generally enhances performance, particularly in Anomaly Detection and Defect Classification tasks, where accuracy significantly improves from 0-shot to 1-shot. This suggests that a single normal sample provides valuable contextual information, enabling the model to make more accurate judgments. However, including more than one reference sample tends to yield diminishing returns and can lead to incorrect judgments, particularly in tasks that benefit more from a straightforward baseline comparison. Notably, larger models such as Qwen2-VL 7B demonstrate superior capabilities in understanding and integrating multiple reference images, consistently outperforming the smaller Qwen2-VL 2B model. Despite this advantage, the use of multiple reference samples may still induce “hallucinations” in the model, highlighting the current limitations of MLLMs in effectively processing and understanding multiple images in complex tasks.

\subsection{Ablation Study}

\begin{table}[t]
    \centering
    \caption{Ablation studies. Here REr denotes Reference Extractor, KG denotes Knowledge Guide, REx denotes Reasoning Expert.}
    \resizebox{\columnwidth}{!}{%
        \begin{tabular}{lccccc}
        \hline
        \textbf{Question Type} & \textbf{Baseline} & \textbf{w/o REr} & \textbf{w/o KG} & \textbf{w/o REx} & \textbf{w/o all} \\ \hline
        Anomaly Detection      & 83.22             & 78.70                            & -                            & -                             & 77.63            \\
        Defect Classification  & 67.61             & 66.65                            & 63.55                        & -                             & 62.76            \\
        Defect Localization    & 69.76             & -                                & -                            & -                             & 69.76            \\
        Defect Description     & 79.80             & 79.72                            & 78.75                        & 79.84                         & 78.45            \\
        Defect Analysis        & 86.16             & 85.05                            & -                            & 84.80                         & 85.05            \\ \hline
        \end{tabular}
        }
  \label{tab:ablation}
\end{table}

In this section, we conduct an ablation study to evaluate the contributions of each expert module in our framework, specifically the Reference Extractor (REr), Knowledge Guide (KG), and Reasoning Expert (REx). \cref{tab:ablation} presents the results across different question types: Anomaly Detection, Defect Classification, Defect Localization, Defect Description, and Defect Analysis. For each task, we report the baseline performance with all modules active, followed by the performance when one or all modules are removed.
The results indicate that each module plays a significant role in enhancing the accuracy of the model. For instance, in Anomaly Detection, removing the Reference Extractor decreases performance from a baseline of 83.22\% to 78.70\%, while removing all modules reduces accuracy further to 77.63\%. This trend is also observed in Defect Classification, where the absence of the Knowledge Guide has a marked impact, reducing accuracy from 67.61\% to 63.55\%. Similarly, in Defect Description, removing the Knowledge Guide and Reasoning Expert individually reduces accuracy, indicating the importance of each module in tasks requiring contextual understanding and reasoning.

\section{Conclusion} \label{sec:conclusion}

In this paper, we introduce a multi-expert framework named Echo, designed to enhance industrial anomaly detection in MLLMs. By focusing on the critical role of visual information in the decision-making process, our work underscores the importance of incorporating visual cues during the decision stage, effectively reducing the hallucination issues that may arise in VLMs. The novelty of the Echo framework also lies in its ability to enable MLLMs to emulate the functions of domain-specific experts, thereby acquiring high-level contextual information and facilitating more rational, in-depth reasoning processes. Demonstrated on the challenging benchmarks of MVTec-AD and VisA, which involve complex industrial anomaly detection tasks, Echo has shown remarkable adaptability and effectiveness, highlighting its strong potential to address real-world challenges in various industrial domains.
{
    \small
    \bibliographystyle{ieeenat_fullname}
    \bibliography{main}

\begin{thebibliography}{41}
\providecommand{\natexlab}[1]{#1}
\providecommand{\url}[1]{\texttt{#1}}
\expandafter\ifx\csname urlstyle\endcsname\relax
  \providecommand{\doi}[1]{doi: #1}\else
  \providecommand{\doi}{doi: \begingroup \urlstyle{rm}\Url}\fi

\bibitem[Bai et~al.(2023)Bai, Bai, Yang, Wang, Tan, Wang, Lin, Zhou, and Zhou]{Qwen-VL}
Jinze Bai, Shuai Bai, Shusheng Yang, Shijie Wang, Sinan Tan, Peng Wang, Junyang Lin, Chang Zhou, and Jingren Zhou.
\newblock Qwen-vl: A versatile vision-language model for understanding, localization, text reading, and beyond.
\newblock \emph{arXiv preprint arXiv:2308.12966}, 2023.

\bibitem[Bergmann et~al.(2019)Bergmann, Fauser, Sattlegger, and Steger]{bergmann2019mvtec}
Paul Bergmann, Michael Fauser, David Sattlegger, and Carsten Steger.
\newblock Mvtec ad--a comprehensive real-world dataset for unsupervised anomaly detection.
\newblock In \emph{Proceedings of the IEEE/CVF conference on computer vision and pattern recognition}, pages 9592--9600, 2019.

\bibitem[Cao et~al.(2024)Cao, Xu, Zhang, Cheng, Huang, Pang, and Shen]{cao2024survey}
Yunkang Cao, Xiaohao Xu, Jiangning Zhang, Yuqi Cheng, Xiaonan Huang, Guansong Pang, and Weiming Shen.
\newblock A survey on visual anomaly detection: Challenge, approach, and prospect.
\newblock \emph{arXiv preprint arXiv:2401.16402}, 2024.

\bibitem[Chen et~al.(2021)Chen, Ding, Zhao, Zhang, Wu, and Shao]{chen2021surface}
Yajun Chen, Yuanyuan Ding, Fan Zhao, Erhu Zhang, Zhangnan Wu, and Linhao Shao.
\newblock Surface defect detection methods for industrial products: A review.
\newblock \emph{Applied Sciences}, 11\penalty0 (16):\penalty0 7657, 2021.

\bibitem[Chen et~al.(2023)Chen, Wu, Wang, Su, Chen, Xing, Zhong, Zhang, Zhu, Lu, Li, Luo, Lu, Qiao, and Dai]{chen2023internvl}
Zhe Chen, Jiannan Wu, Wenhai Wang, Weijie Su, Guo Chen, Sen Xing, Muyan Zhong, Qinglong Zhang, Xizhou Zhu, Lewei Lu, Bin Li, Ping Luo, Tong Lu, Yu Qiao, and Jifeng Dai.
\newblock Internvl: Scaling up vision foundation models and aligning for generic visual-linguistic tasks.
\newblock \emph{arXiv preprint arXiv:2312.14238}, 2023.

\bibitem[Chen et~al.(2024{\natexlab{a}})Chen, Wang, Tian, Ye, Gao, Cui, Tong, Hu, Luo, Ma, et~al.]{chen2024far}
Zhe Chen, Weiyun Wang, Hao Tian, Shenglong Ye, Zhangwei Gao, Erfei Cui, Wenwen Tong, Kongzhi Hu, Jiapeng Luo, Zheng Ma, et~al.
\newblock How far are we to gpt-4v? closing the gap to commercial multimodal models with open-source suites.
\newblock \emph{arXiv preprint arXiv:2404.16821}, 2024{\natexlab{a}}.

\bibitem[Chen et~al.(2024{\natexlab{b}})Chen, Xu, Qi, and Guo]{chen2024mllm}
Zhanpeng Chen, Chengjin Xu, Yiyan Qi, and Jian Guo.
\newblock Mllm is a strong reranker: Advancing multimodal retrieval-augmented generation via knowledge-enhanced reranking and noise-injected training.
\newblock \emph{arXiv preprint arXiv:2407.21439}, 2024{\natexlab{b}}.

\bibitem[Deng et~al.(2024)Deng, Luo, Zhai, Cao, and Kang]{deng2024vmad}
Huilin Deng, Hongchen Luo, Wei Zhai, Yang Cao, and Yu Kang.
\newblock Vmad: Visual-enhanced multimodal large language model for zero-shot anomaly detection.
\newblock \emph{arXiv preprint arXiv:2409.20146}, 2024.

\bibitem[Gao et~al.(2024)Gao, Chen, Zhang, Fu, Shen, Zhang, Zhang, Zheng, Sun, Cao, et~al.]{gao2024cantor}
Timin Gao, Peixian Chen, Mengdan Zhang, Chaoyou Fu, Yunhang Shen, Yan Zhang, Shengchuan Zhang, Xiawu Zheng, Xing Sun, Liujuan Cao, et~al.
\newblock Cantor: Inspiring multimodal chain-of-thought of mllm.
\newblock In \emph{Proceedings of the 32nd ACM International Conference on Multimedia}, pages 9096--9105, 2024.

\bibitem[Gu et~al.(2024)Gu, Zhu, Zhu, Chen, Tang, and Wang]{gu2024anomalygpt}
Zhaopeng Gu, Bingke Zhu, Guibo Zhu, Yingying Chen, Ming Tang, and Jinqiao Wang.
\newblock Anomalygpt: Detecting industrial anomalies using large vision-language models.
\newblock In \emph{Proceedings of the AAAI Conference on Artificial Intelligence}, pages 1932--1940, 2024.

\bibitem[Hu et~al.(2023)Hu, Iscen, Sun, Wang, Chang, Sun, Schmid, Ross, and Fathi]{hu2023reveal}
Ziniu Hu, Ahmet Iscen, Chen Sun, Zirui Wang, Kai-Wei Chang, Yizhou Sun, Cordelia Schmid, David~A Ross, and Alireza Fathi.
\newblock Reveal: Retrieval-augmented visual-language pre-training with multi-source multimodal knowledge memory.
\newblock In \emph{Proceedings of the IEEE/CVF conference on computer vision and pattern recognition}, pages 23369--23379, 2023.

\bibitem[Jeong et~al.(2023)Jeong, Zou, Kim, Zhang, Ravichandran, and Dabeer]{jeong2023winclip}
Jongheon Jeong, Yang Zou, Taewan Kim, Dongqing Zhang, Avinash Ravichandran, and Onkar Dabeer.
\newblock Winclip: Zero-/few-shot anomaly classification and segmentation.
\newblock In \emph{Proceedings of the IEEE/CVF Conference on Computer Vision and Pattern Recognition}, pages 19606--19616, 2023.

\bibitem[Jiang et~al.(2024)Jiang, Li, Deng, Liu, Gao, Zhou, Li, Wang, and Zheng]{jiang2024mmad}
Xi Jiang, Jian Li, Hanqiu Deng, Yong Liu, Bin-Bin Gao, Yifeng Zhou, Jialin Li, Chengjie Wang, and Feng Zheng.
\newblock Mmad: The first-ever comprehensive benchmark for multimodal large language models in industrial anomaly detection.
\newblock \emph{arXiv preprint arXiv:2410.09453}, 2024.

\bibitem[Johnson et~al.(2019)Johnson, Douze, and J{\'e}gou]{johnson2019billion}
Jeff Johnson, Matthijs Douze, and Herv{\'e} J{\'e}gou.
\newblock Billion-scale similarity search with {GPUs}.
\newblock \emph{IEEE Transactions on Big Data}, 7\penalty0 (3):\penalty0 535--547, 2019.

\bibitem[Lewis et~al.(2020)Lewis, Perez, Piktus, Petroni, Karpukhin, Goyal, K{\"u}ttler, Lewis, Yih, Rockt{\"a}schel, et~al.]{lewis2020retrieval}
Patrick Lewis, Ethan Perez, Aleksandra Piktus, Fabio Petroni, Vladimir Karpukhin, Naman Goyal, Heinrich K{\"u}ttler, Mike Lewis, Wen-tau Yih, Tim Rockt{\"a}schel, et~al.
\newblock Retrieval-augmented generation for knowledge-intensive nlp tasks.
\newblock \emph{Advances in Neural Information Processing Systems}, 33:\penalty0 9459--9474, 2020.

\bibitem[Li et~al.(2024{\natexlab{a}})Li, Zhang, Guo, Zhang, Li, Zhang, Zhang, Zhang, Li, Liu, and Li]{li2024llavaonevisioneasyvisualtask}
Bo Li, Yuanhan Zhang, Dong Guo, Renrui Zhang, Feng Li, Hao Zhang, Kaichen Zhang, Peiyuan Zhang, Yanwei Li, Ziwei Liu, and Chunyuan Li.
\newblock Llava-onevision: Easy visual task transfer, 2024{\natexlab{a}}.

\bibitem[Li et~al.(2024{\natexlab{b}})Li, Zhang, Zhang, Zhang, Li, Li, Ma, and Li]{li2024llavanextinterleavetacklingmultiimagevideo}
Feng Li, Renrui Zhang, Hao Zhang, Yuanhan Zhang, Bo Li, Wei Li, Zejun Ma, and Chunyuan Li.
\newblock Llava-next-interleave: Tackling multi-image, video, and 3d in large multimodal models, 2024{\natexlab{b}}.

\bibitem[Li et~al.(2023)Li, Wang, Yuan, Liu, Zhao, Guo, Xu, Shi, and Zuo]{li2023myriad}
Yuanze Li, Haolin Wang, Shihao Yuan, Ming Liu, Debin Zhao, Yiwen Guo, Chen Xu, Guangming Shi, and Wangmeng Zuo.
\newblock Myriad: Large multimodal model by applying vision experts for industrial anomaly detection.
\newblock \emph{arXiv preprint arXiv:2310.19070}, 2023.

\bibitem[Li et~al.(2024{\natexlab{c}})Li, Goodge, Liu, and Foo]{li2024promptad}
Yiting Li, Adam Goodge, Fayao Liu, and Chuan-Sheng Foo.
\newblock Promptad: Zero-shot anomaly detection using text prompts.
\newblock In \emph{Proceedings of the IEEE/CVF Winter Conference on Applications of Computer Vision}, pages 1093--1102, 2024{\natexlab{c}}.

\bibitem[Liu et~al.(2023{\natexlab{a}})Liu, Li, Li, and Lee]{liu2023improvedllava}
Haotian Liu, Chunyuan Li, Yuheng Li, and Yong~Jae Lee.
\newblock Improved baselines with visual instruction tuning, 2023{\natexlab{a}}.

\bibitem[Liu et~al.(2023{\natexlab{b}})Liu, Li, Wu, and Lee]{liu2023llava}
Haotian Liu, Chunyuan Li, Qingyang Wu, and Yong~Jae Lee.
\newblock Visual instruction tuning.
\newblock In \emph{NeurIPS}, 2023{\natexlab{b}}.

\bibitem[Liu et~al.(2023{\natexlab{c}})Liu, Son, Yang, Liu, Gao, Lee, and Li]{liu2023learning}
Haotian Liu, Kilho Son, Jianwei Yang, Ce Liu, Jianfeng Gao, Yong~Jae Lee, and Chunyuan Li.
\newblock Learning customized visual models with retrieval-augmented knowledge.
\newblock In \emph{Proceedings of the IEEE/CVF Conference on Computer Vision and Pattern Recognition}, pages 15148--15158, 2023{\natexlab{c}}.

\bibitem[Liu et~al.(2024{\natexlab{a}})Liu, Xie, Wang, Li, Wang, Zheng, and Jin]{liu2024deep}
Jiaqi Liu, Guoyang Xie, Jinbao Wang, Shangnian Li, Chengjie Wang, Feng Zheng, and Yaochu Jin.
\newblock Deep industrial image anomaly detection: A survey.
\newblock \emph{Machine Intelligence Research}, 21\penalty0 (1):\penalty0 104--135, 2024{\natexlab{a}}.

\bibitem[Liu et~al.(2024{\natexlab{b}})Liu, Sun, Zang, Li, Zhang, Dong, Xiong, Lin, and Wang]{liu2024rar}
Ziyu Liu, Zeyi Sun, Yuhang Zang, Wei Li, Pan Zhang, Xiaoyi Dong, Yuanjun Xiong, Dahua Lin, and Jiaqi Wang.
\newblock Rar: Retrieving and ranking augmented mllms for visual recognition.
\newblock \emph{arXiv preprint arXiv:2403.13805}, 2024{\natexlab{b}}.

\bibitem[Malkov and Yashunin(2018{\natexlab{a}})]{malkov2018efficient}
Yu~A Malkov and Dmitry~A Yashunin.
\newblock Efficient and robust approximate nearest neighbor search using hierarchical navigable small world graphs.
\newblock \emph{IEEE transactions on pattern analysis and machine intelligence}, 42\penalty0 (4):\penalty0 824--836, 2018{\natexlab{a}}.

\bibitem[Malkov and Yashunin(2018{\natexlab{b}})]{malkov2018efficientrobustapproximatenearest}
Yu.~A. Malkov and D.~A. Yashunin.
\newblock Efficient and robust approximate nearest neighbor search using hierarchical navigable small world graphs, 2018{\natexlab{b}}.

\bibitem[Radford et~al.(2021)Radford, Kim, Hallacy, Ramesh, Goh, Agarwal, Sastry, Askell, Mishkin, Clark, et~al.]{radford2021learning}
Alec Radford, Jong~Wook Kim, Chris Hallacy, Aditya Ramesh, Gabriel Goh, Sandhini Agarwal, Girish Sastry, Amanda Askell, Pamela Mishkin, Jack Clark, et~al.
\newblock Learning transferable visual models from natural language supervision.
\newblock In \emph{International conference on machine learning}, pages 8748--8763. PMLR, 2021.

\bibitem[Rudolph et~al.(2022)Rudolph, Wehrbein, Rosenhahn, and Wandt]{rudolph2022fully}
Marco Rudolph, Tom Wehrbein, Bodo Rosenhahn, and Bastian Wandt.
\newblock Fully convolutional cross-scale-flows for image-based defect detection.
\newblock In \emph{Proceedings of the IEEE/CVF Winter Conference on Applications of Computer Vision}, pages 1088--1097, 2022.

\bibitem[Tao et~al.(2022)Tao, Gong, Zhang, Yan, and Adak]{tao2022deep}
Xian Tao, Xinyi Gong, Xin Zhang, Shaohua Yan, and Chandranath Adak.
\newblock Deep learning for unsupervised anomaly localization in industrial images: A survey.
\newblock \emph{IEEE Transactions on Instrumentation and Measurement}, 71:\penalty0 1--21, 2022.

\bibitem[Tax and Duin(2004)]{tax2004support}
David~MJ Tax and Robert~PW Duin.
\newblock Support vector data description.
\newblock \emph{Machine learning}, 54:\penalty0 45--66, 2004.

\bibitem[Wang et~al.(2024)Wang, Bai, Tan, Wang, Fan, Bai, Chen, Liu, Wang, Ge, Fan, Dang, Du, Ren, Men, Liu, Zhou, Zhou, and Lin]{Qwen2VL}
Peng Wang, Shuai Bai, Sinan Tan, Shijie Wang, Zhihao Fan, Jinze Bai, Keqin Chen, Xuejing Liu, Jialin Wang, Wenbin Ge, Yang Fan, Kai Dang, Mengfei Du, Xuancheng Ren, Rui Men, Dayiheng Liu, Chang Zhou, Jingren Zhou, and Junyang Lin.
\newblock Qwen2-vl: Enhancing vision-language model's perception of the world at any resolution.
\newblock \emph{arXiv preprint arXiv:2409.12191}, 2024.

\bibitem[Wei et~al.(2022)Wei, Wang, Schuurmans, Bosma, Xia, Chi, Le, Zhou, et~al.]{wei2022chain}
Jason Wei, Xuezhi Wang, Dale Schuurmans, Maarten Bosma, Fei Xia, Ed Chi, Quoc~V Le, Denny Zhou, et~al.
\newblock Chain-of-thought prompting elicits reasoning in large language models.
\newblock \emph{Advances in neural information processing systems}, 35:\penalty0 24824--24837, 2022.

\bibitem[Wei et~al.(2023)Wei, Huang, Lu, Zhou, and Le]{wei2023simple}
Jerry Wei, Da Huang, Yifeng Lu, Denny Zhou, and Quoc~V Le.
\newblock Simple synthetic data reduces sycophancy in large language models.
\newblock \emph{arXiv preprint arXiv:2308.03958}, 2023.

\bibitem[Yao et~al.(2024)Yao, Yu, Zhang, Wang, Cui, Zhu, Cai, Li, Zhao, He, et~al.]{yao2024minicpm}
Yuan Yao, Tianyu Yu, Ao Zhang, Chongyi Wang, Junbo Cui, Hongji Zhu, Tianchi Cai, Haoyu Li, Weilin Zhao, Zhihui He, et~al.
\newblock Minicpm-v: A gpt-4v level mllm on your phone.
\newblock \emph{arXiv preprint arXiv:2408.01800}, 2024.

\bibitem[Yi and Yoon(2020)]{yi2020patch}
Jihun Yi and Sungroh Yoon.
\newblock Patch svdd: Patch-level svdd for anomaly detection and segmentation.
\newblock In \emph{Proceedings of the Asian conference on computer vision}, 2020.

\bibitem[Yuan et~al.(2024)Yuan, Sun, Omeiza, Zhao, Newman, Kunze, and Gadd]{yuan2024rag}
Jianhao Yuan, Shuyang Sun, Daniel Omeiza, Bo Zhao, Paul Newman, Lars Kunze, and Matthew Gadd.
\newblock Rag-driver: Generalisable driving explanations with retrieval-augmented in-context learning in multi-modal large language model.
\newblock \emph{arXiv preprint arXiv:2402.10828}, 2024.

\bibitem[Zavrtanik et~al.(2021)Zavrtanik, Kristan, and Sko{\v{c}}aj]{zavrtanik2021reconstruction}
Vitjan Zavrtanik, Matej Kristan, and Danijel Sko{\v{c}}aj.
\newblock Reconstruction by inpainting for visual anomaly detection.
\newblock \emph{Pattern Recognition}, 112:\penalty0 107706, 2021.

\bibitem[Zhang et~al.(2023)Zhang, Zhang, Li, Zhao, Karypis, and Smola]{zhang2023multimodal}
Zhuosheng Zhang, Aston Zhang, Mu Li, Hai Zhao, George Karypis, and Alex Smola.
\newblock Multimodal chain-of-thought reasoning in language models.
\newblock \emph{arXiv preprint arXiv:2302.00923}, 2023.

\bibitem[Zhao et~al.(2023)Zhao, Chen, Wang, Jiao, Do, Qin, Ding, Guo, Li, Li, et~al.]{zhao2023retrieving}
Ruochen Zhao, Hailin Chen, Weishi Wang, Fangkai Jiao, Xuan~Long Do, Chengwei Qin, Bosheng Ding, Xiaobao Guo, Minzhi Li, Xingxuan Li, et~al.
\newblock Retrieving multimodal information for augmented generation: A survey.
\newblock \emph{arXiv preprint arXiv:2303.10868}, 2023.

\bibitem[Zhou et~al.(2023)Zhou, Pang, Tian, He, and Chen]{zhou2023anomalyclip}
Qihang Zhou, Guansong Pang, Yu Tian, Shibo He, and Jiming Chen.
\newblock Anomalyclip: Object-agnostic prompt learning for zero-shot anomaly detection.
\newblock \emph{arXiv preprint arXiv:2310.18961}, 2023.

\bibitem[Zou et~al.(2022)Zou, Jeong, Pemula, Zhang, and Dabeer]{zou2022spot}
Yang Zou, Jongheon Jeong, Latha Pemula, Dongqing Zhang, and Onkar Dabeer.
\newblock Spot-the-difference self-supervised pre-training for anomaly detection and segmentation.
\newblock In \emph{European Conference on Computer Vision}, pages 392--408. Springer, 2022.

\end{thebibliography}
}

\clearpage
\setcounter{page}{1}
\renewcommand{\thesection}{\Alph{section}}
\setcounter{section}{0}
\maketitlesupplementary

\section{Additional Analysis}

\subsection{Case Presentation}

\begin{figure*}[]
  \centering
   \includegraphics[width=0.66\linewidth]{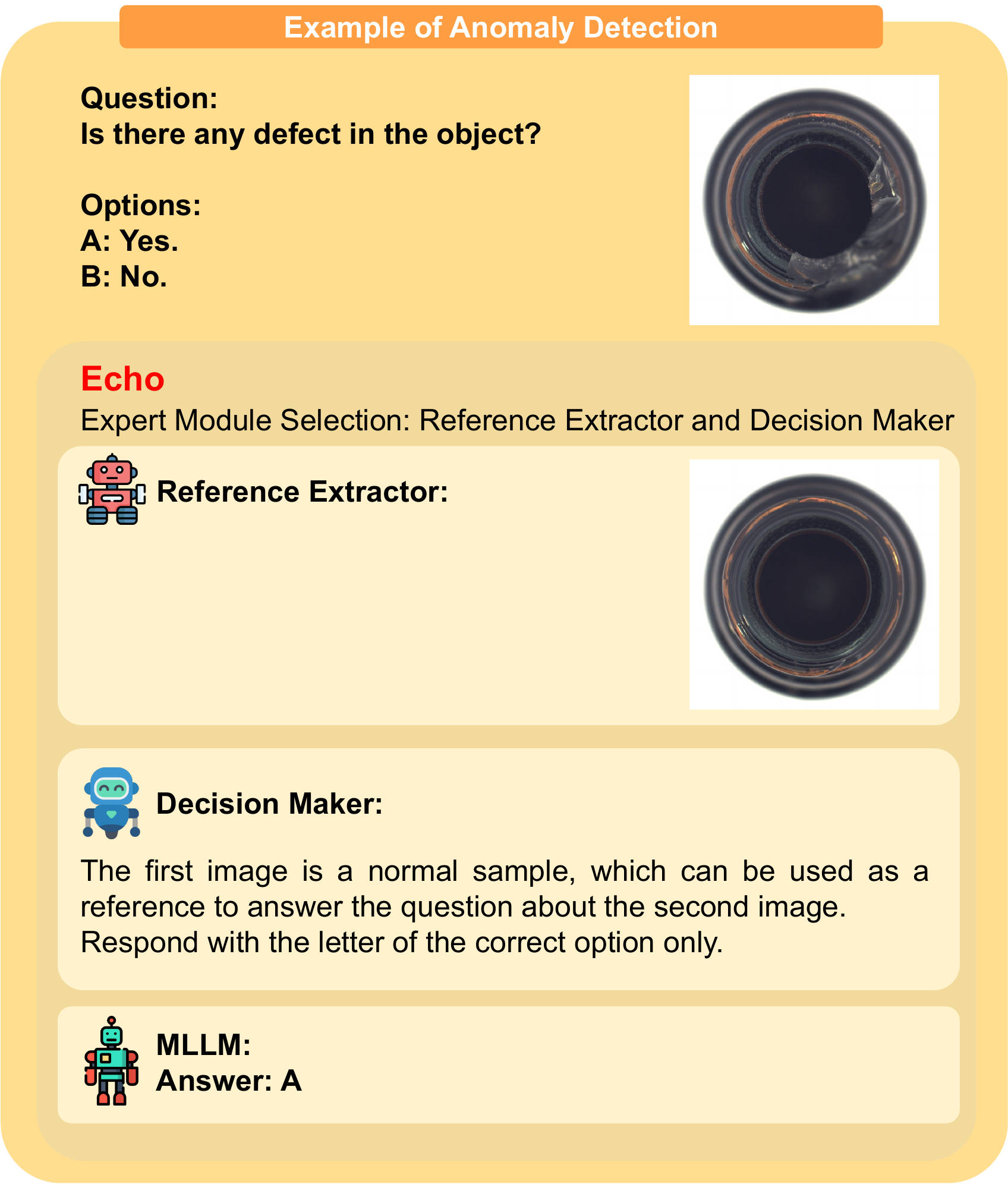}
   \caption{Example of Anomaly Detection.}
   \label{Anomaly_Detection}
\end{figure*}

\begin{figure*}[]
  \centering
   \includegraphics[width=0.66\linewidth]{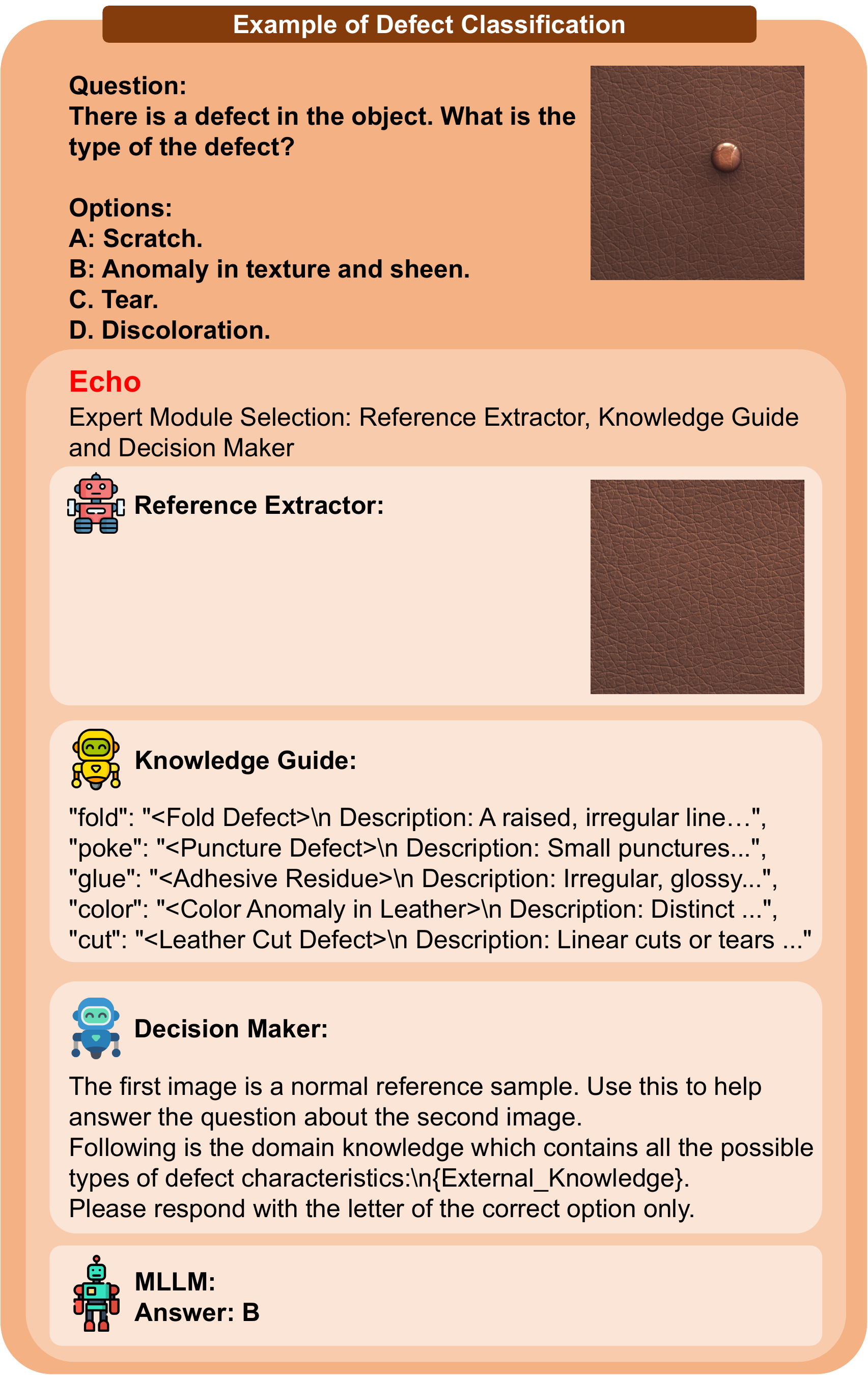}
   \caption{Example of Defect Classification.}
   \label{Defect_Classification}
\end{figure*}

\begin{figure*}[]
  \centering
   \includegraphics[width=0.66\linewidth]{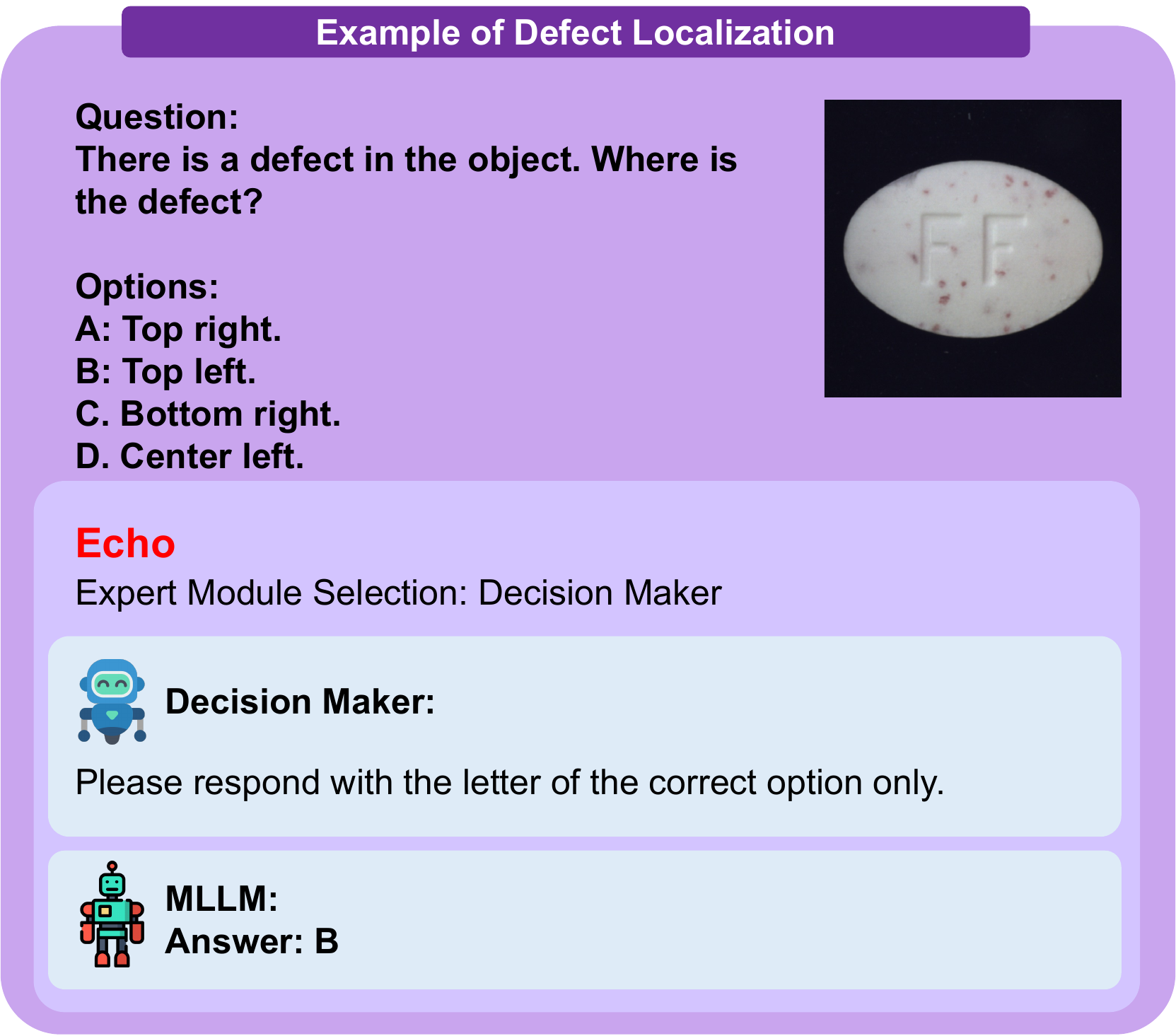}
   \caption{Example of Defect Localization.}
   \label{Defect_Localization}
\end{figure*}

\begin{figure*}[]
  \centering
   \includegraphics[width=0.66\linewidth]{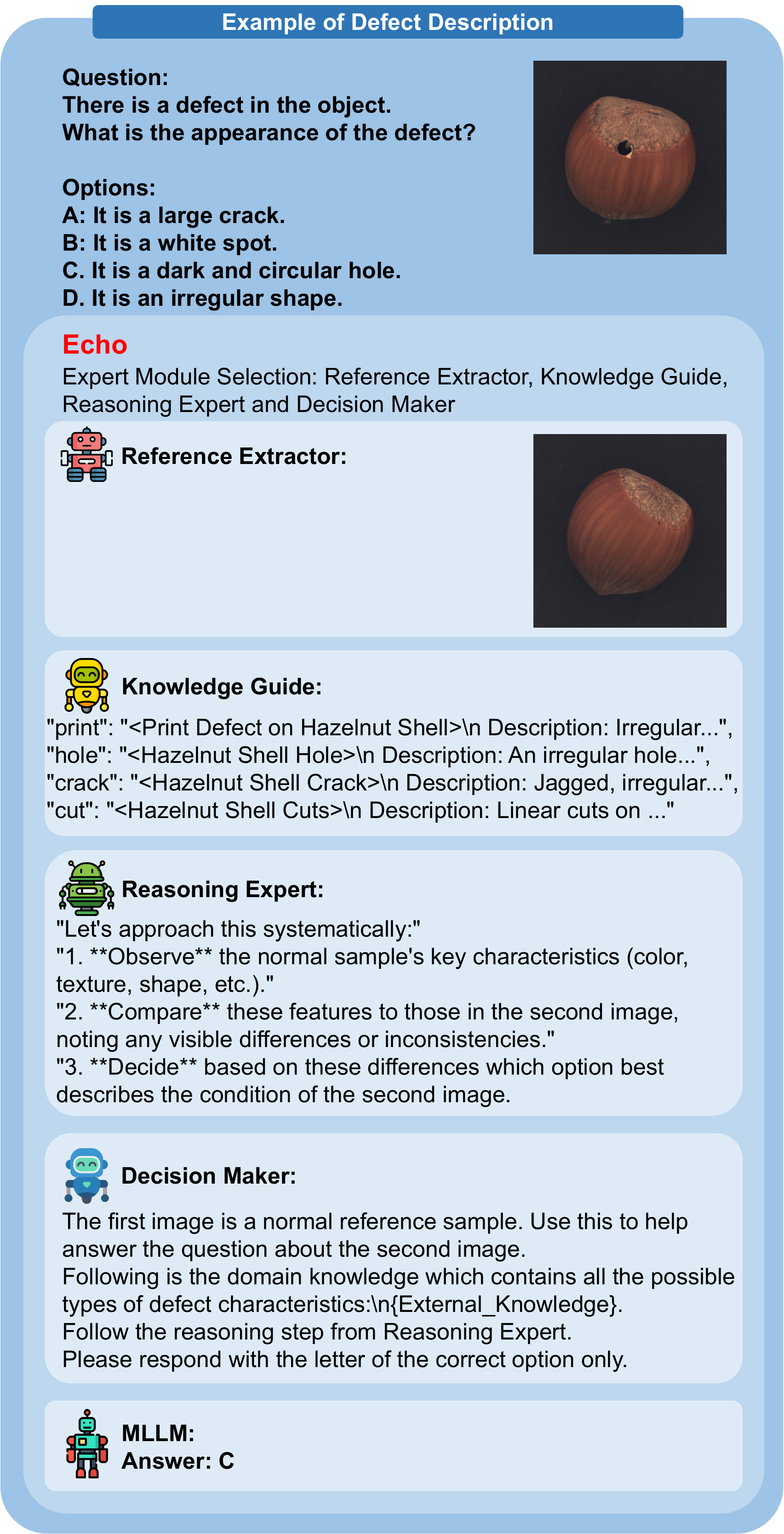}
   \caption{Example of Defect Description.}
   \label{Defect_Description}
\end{figure*}

\begin{figure*}[]
  \centering
   \includegraphics[width=0.66\linewidth]{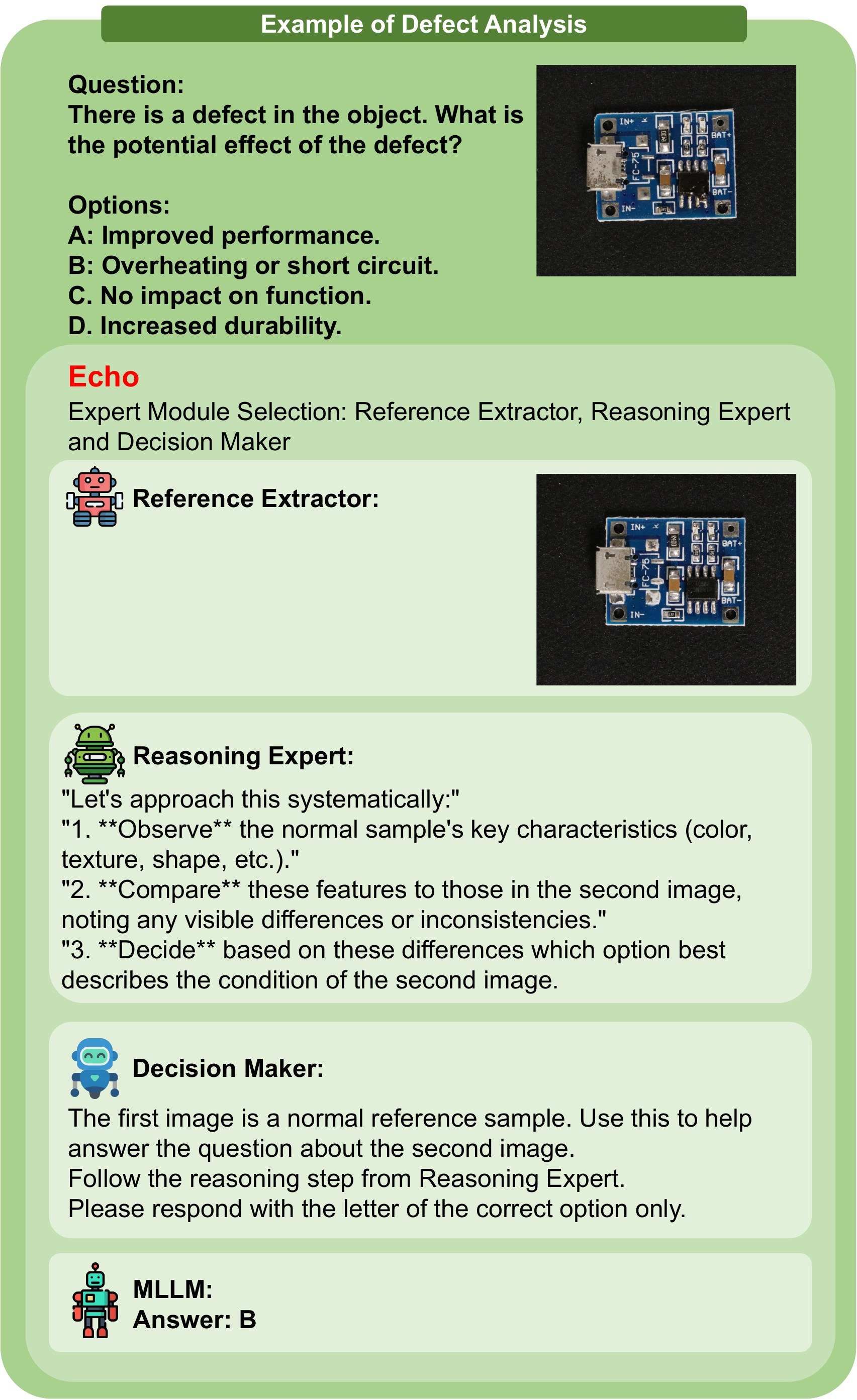}
   \caption{Example of Defect Analysis.}
   \label{Defect_Analysis}
\end{figure*}

In this section, we delve into the detailed design of the prompts used in Echo. The prompts are tailored for various industrial anomaly detection tasks and aim to maximize the efficiency and precision of MLLMs in generating context-aware decisions. Echo incorporates a structured, multi-expert pipeline, and the prompts are a critical component of this system, guiding the reasoning and decision-making processes across different modules. In Figs \ref{Anomaly_Detection}, \ref{Defect_Classification}, \ref{Defect_Localization}, \ref{Defect_Description}, \ref{Defect_Analysis}, we show some specific cases of Echo. It can be seen that Echo has good decision-generating and practical problem-solving abilities in IAD tasks.

\subsection{Impact of Multimodal Retriever}
In this section, we evaluate the performance of the Echo framework under the 0-shot, 1-shot, and 1-shot* settings to analyze the impact of integrating multimodal retrieval into the \textit{Reference Extractor} expert module, as shown in \cref{exp: reference}. In the 0-shot setting, the model processes only the query image using the \textit{Knowledge Guide}, \textit{Reasoning Expert}, and \textit{Decision Maker} expert modules without relying on a reference image. In the 1-shot setting, a randomly selected normal image of the same type object as the query image is provided as a reference, allowing the model to incorporate general contextual information. In contrast, the 1-shot* setting leverages the multimodal retriever in the \textit{Reference Extractor} to retrieve the most similar normal image. 

As shown in \cref{exp: reference}, the results reveal a clear distinction between the performance of the Qwen2-VL 2B and Qwen2-VL 7B models under different settings, shedding light on the varying capabilities of smaller and larger models in handling multi-image contextualization. For Qwen2-VL 2B, the 1-shot setting provides a slight improvement in anomaly detection (+0.67\%) compared to 0-shot, but for other tasks such as defect classification and defect description, the performance drops below the 0-shot baseline. Similarly, the 1-shot* setting yields only marginal improvements in anomaly detection (+0.68\%), while the results for defect classification and description remain slightly worse than 0-shot. These findings suggest that smaller models like Qwen2-VL 2B struggle to effectively utilize multi-image information, showing their limitations in reasoning across multiple visual inputs.

In contrast, Qwen2-VL 7B demonstrates significant improvements across all tasks in both 1-shot and 1-shot* settings, with the gains being especially pronounced in the 1-shot* setting. For anomaly detection, the 1-shot* setting achieves a remarkable +5.99\% improvement over 0-shot, and the overall mean accuracy increases by +3.23\%. These results indicate that larger models, such as Qwen2-VL 7B, possess a stronger ability to understand and reason about multi-image contexts, leveraging the additional reference image to make more accurate predictions. Furthermore, the substantial difference in performance between the 1-shot and 1-shot* settings for Qwen2-VL 7B underscores the importance of Echo’s \textit{Reference Extractor}. By retrieving the most similar normal image, the \textit{Reference Extractor} enables the model to effectively compare and contextualize visual inputs, significantly enhancing its anomaly detection, classification, and reasoning capabilities. This demonstrates that the advanced retrieval mechanism is crucial for fully utilizing the potential of large models, while smaller models still require further optimization to handle multi-image inputs effectively.

\begin{table}[h!]
\centering
\caption{The impact of reference images. (0-shot indicates no reference image, 1-shot indicates using a random reference image, 1-shot* indicates using a retrieved most similar image as reference visual information.)}
\resizebox{\columnwidth}{!}{%
\begin{tblr}{
  column{3} = {c},
  column{4} = {c},
  column{5} = {c},
  column{6} = {c},
  column{7} = {c},
  cell{2}{1} = {r=3}{},
  cell{5}{1} = {r=3}{},
  hline{1-2,5,8} = {-}{},
}
\textbf{Model} & \textbf{Setting} & \textbf{Anomaly Detection}              & \textbf{Defect Classification}          & \textbf{Defect Description}             & \textbf{Defect Analysis}                & \textbf{Mean}                           \\
Qwen2-VL 2B    & 0 shot           & 66.50                                   & \textbf{48.70}                          & \textbf{70.84}                          & \textbf{78.58}                          & \textbf{66.16}                          \\
               & 1 shot           & 67.17 (\textcolor{red}{+0.67})          & 46.95 (\textcolor{green}{-1.75})        & 70.11 (\textcolor{green}{-0.73})        & 77.02 (\textcolor{green}{-1.56})        & 65.31 (\textcolor{green}{-0.85})        \\
               & 1 shot *         & \textbf{67.18} (\textcolor{red}{+0.68}) & 46.87 (\textcolor{green}{-1.83})        & 70.07 (\textcolor{green}{-0.77})        & 77.24 (\textcolor{green}{-1.34})        & 65.34 (\textcolor{green}{-0.82})        \\
Qwen2-VL 7B    & 0 shot           & 77.63                                   & 62.76                                   & 78.45                                   & 85.05                                   & 75.97                                   \\
               & 1 shot           & 82.62 (\textcolor{red}{+4.99})          & 67.12 (\textcolor{red}{+4.36})          & 79.18 (\textcolor{red}{+0.73})          & \textbf{86.24} (\textcolor{red}{+1.19}) & 78.79 (\textcolor{red}{+2.82})          \\
               & 1 shot *         & \textbf{83.22} (\textcolor{red}{+5.99}) & \textbf{67.61} (\textcolor{red}{+4.85}) & \textbf{79.80} (\textcolor{red}{+1.35}) & 86.16 (\textcolor{red}{+1.11})          & \textbf{79.20} (\textcolor{red}{+3.23}) 
\end{tblr}
}
\label{exp: reference}
\end{table}

\subsection{Limitations}
Although the datasets used in this paper are high-quality, images of fine industrial components in real-world factories often suffer from low resolution, noise, or poor lighting. However, the ability of MLLMs to handle such low-quality images remains unexplored and requires further study. Furthermore, in industrial settings where computational resources are limited, especially at the edge, it is crucial to focus on smaller models like those in the 2B parameter range to make the framework practical. Moreover, while the current evaluation is based on datasets with predefined defect information, the ability of the framework and MLLMs to detect or reason about novel defect types remains an open question. This is particularly important for tasks like defect classification, localization, description, and analysis, which require further research to enhance robustness and adaptability.

\end{document}